%% file: iclr2020_conference.tex
\documentclass{article} 
\usepackage{iclr2020_conference,times}

\input{math_commands.tex}

\usepackage{graphicx}
\usepackage{subcaption}
\usepackage{hyperref}
\usepackage{url}

\title{Importance of using appropriate baselines for evaluation of data-efficiency in deep reinforcement learning for Atari}

\author{Kacper P. Kielak \\
School of Computer Science\\
University of Birmingham\\
Birmingham, B15 2TT, UK \\
\texttt{k.kielak@bham.ac.uk} \\
}

\iclrfinalcopy 
\begin{document}

\maketitle 

\begin{abstract}
Reinforcement learning (RL) has seen great advancements in the past few years. Nevertheless, the consensus among the RL community is that currently used methods, despite all their benefits, suffer from extreme data inefficiency, especially in the rich visual domains like Atari. To circumvent this problem, novel approaches were introduced that often claim to be much more efficient than popular variations of the state-of-the-art DQN algorithm. In this paper, however, we demonstrate that the newly proposed techniques simply used unfair baselines in their experiments. Namely, we show that the actual improvement in the efficiency came from allowing the algorithm for more training updates for each data sample, and not from employing the new methods. By allowing DQN to execute network updates more frequently we manage to reach similar or better results than the recently proposed advancement, often at a fraction of complexity and computational costs. Furthermore, based on the outcomes of the study, we argue that the agent similar to the modified DQN that is presented in this paper should be used as a baseline for any future work aimed at improving sample efficiency of deep reinforcement learning.
\end{abstract}

\section{Introduction}

Producing fully independent agents that learn optimal behavior and develop over time purely by trial and error interaction with the surrounding environment is one of the prominent dilemmas in the field of artificial intelligence. A mathematical framework that encapsulates the problem of these autonomous systems is reinforcement learning. Over the past few years, exceptional progress has been made in devising artificial agents that can learn and solve problems in a variety of domains using deep RL methods \citep{mnih2015dqn, schulman2015trpo, silver2016alphago}. However, these algorithms are perceived as extremely data inefficient. They are thought to require an immense amount of non-optimal interaction with the real environment before they begin to operate acceptably well \citep{irpan2018doesntwork}. 

One of the most popular benchmarks for assessing overall performance and data complexity of deep RL algorithms is Atari Learning Environment \citep{bellemare2013ale, machado2018ale}. The state-of-the-art approaches, at least in the way they were presented so far, need millions of frames to learn how to play these games acceptably well \citep{schulman2017ppo, hessel2018rainbow}. It corresponds to days of play experience using the standard frame rate. However, human players can achieve the same within minutes \citep{tsividis2017human}.

A lot of work has been produced to circumvent these shortcomings. Most studies focused on the visually-rich Atari domain employ model-based strategies inspired by the classical Dyna approach \citep{sutton1991dyna} and action-conditional prediction methods \citep{oh2015action, leibfried2016reward}. Although some of them manage to drastically reduce the amount of data required by the standard algorithms, they do so by highly increasing both conceptual and computational complexity of the models.

In this paper, we argue, and experimentally prove, that already existing techniques can be much more data-efficient than it is assumed. We introduce a simple change to the DQN-based algorithms. In some environments like Pong or Hero, it can achieve the same results given only 5\% - 10\% of the data it is often presented to need. Furthermore, it results in the same data-efficiency as the recent advancements in the field while often being much more stable, simpler, and requiring much less computation.

Following the introduction, section \ref{sect:background} gives a brief background behind reinforcement learning with the focus on Q-learning and its deep learning equivalents. Section \ref{sect:advancements} provides an overview of recent studies aimed at improving data efficiency. Section \ref{sect:eff} argues that standard DQN-like algorithms can be much more efficient than it tends to be presented and that recently proposed techniques only give an illusion of efficiency. Then, the description and analysis of experiments follow in sections 5 and 6. Finally, section 7 concludes this study.

\section{Background} \label{sect:background}
Reinforcement learning is a problem of learning a policy that maximises the reward signal for a given task. To define RL setting we need a set of possible environment states $\displaystyle \sS$, a set of available actions $\displaystyle \sA$, and relations between those. These relations are described by a transition function $\displaystyle T : \sS \times \sA \rightarrow \sS$ that defines dynamics of transitions from one state to another, and a reward function $\displaystyle R : \sS \times \sA \rightarrow \sR$ that defines the real-valued reward signal. Together, $\displaystyle T$ and $\displaystyle R$ constitute the model of the environment. The goal of reinforcement learning is to find a policy $\displaystyle \pi : \sS \rightarrow \sA$ that maximises the total cumulative reward over time. 
One of the most popular reinforcement learning algorithms is Q-learning \citep{watkins1992qlearning}. Q-learning decides on an optimal policy based on the state-action value function $\displaystyle Q: \sS \times \sA \rightarrow \sR$ that maps state and action performed in that state to the expected total cumulative reward following the action. The algorithm chooses an action that maximizes $\displaystyle Q$, i.e. $\displaystyle a_t = \text{argmax}_a Q(s_t, a)$. $\displaystyle Q$ is learned in the process of interacting with the environment. At every agent's step, tuple $(\displaystyle s_t \in \sS, a_t \in \sA, r_t \in \sR, s_{t+1} \in \sS)$ is obtained and immediately used to update the Q function. Because state-action combination is often too big or continuous to represent directly in a tabular manner, $\displaystyle Q$ is commonly approximated using different supervised learning algorithms. However, using deep learning to approximate $\displaystyle Q$ is not trivial because Q-learning breaks important assumptions required by neural networks. Namely, $\displaystyle Q$ update is recursive and experience tuples are highly correlated when used sequentially.

Recently introduced DQN \citep{mnih2015dqn} bypassed this issue by introducing two concepts: target network and replay buffer. Target network is simply a fixed snapshot of the network that approximates $\displaystyle Q$ value (online network) taken every $\tau_t$ steps. Instead of updating the online network towards itself, it is updated towards the target network.  This approach maintains the logic of Q-learning while stopping the online network from diverging due to recursive updates. Replay buffer, on the other hand, guarantees a much higher level of independence between experience tuples. They are not used immediately, one after another anymore but stored in the replay buffer instead. Then, every $\tau_u$ steps, a single training step is performed, i.e. a mini-batch of randomly sampled experience from the replay buffer is used to update the online network. It reduces the correlation between experience samples by breaking their ordering.

Rainbow DQN \citep{hessel2018rainbow} is a combination of several incremental improvements on top of DQN that increased both sample efficiency and the total performance of the algorithm achieving state-of-the-art results. It is an architecture that we use as an example that current model-free deep RL is not as inefficient as it is often stated. Throughout the paper hyperparameters from \citet{hessel2018rainbow} are employed, unless stated otherwise.

\section{Advancements in data efficiency of reinforcement learning} \label{sect:advancements}

The most promising approach to improving data efficiency of deep RL is based on the premise of model-based techniques \citep{sutton2018intro}. Having access to transition and reward mechanics of the environment would make it possible to construct an artificial simulation where the agent could be trained without performing often costly interactions with the real environment. However, in most scenarios, the agent is not given any prior information about the model of its environment. This issue is often overcome by learning the model instead. \citet{oh2015action} and \citet{leibfried2016reward} have shown that it is possible with a very high level of accuracy. 

Ability to learn the model of the environment was subsequently leveraged to successfully improve different aspects of deep RL \citep{racaniere2017i2a, oh2017value, buesing2018search, ha2018models}. \citet{azizzadenesheli2018gats}, \citet{holland2018dynadqn}, and \citet{kaiser2019model}, however, focused directly on employing the learned models to increase data efficiency of deep RL algorithms. 

\citet{azizzadenesheli2018gats} proposed Generative Adversarial Tree Search (GATS). Unlike in the standard approach to learning the environment dynamics, GATS creates two separate models: Generative Dynamics Model (GDM) based on a modified Pix2Pix \citep{isola2017image} to learn the transition model $\displaystyle T: \sS \times \sA \rightarrow \sS$; and Reward Predictor (RP), a simple 3-class classification architecture to learn the reward model $\displaystyle R: \sS \times \sA \rightarrow \sR$. Both models learn from experience stored in DQN's replay buffer and are then used for bounded Monte Carlo tree search as in \citep{silver2016alphago}. GATS is evaluated primarily on the game Pong where it learns an optimal policy using around 42\% of the data required by using standalone model-free agent what is a tiny improvement compared to the methods described next.

\citet{holland2018dynadqn} explored the performance of the model-based approach given either perfect model, model pretrained on expert data (pretrained model), or model learned alongside the agent's value function (online model). Both non-perfect models followed standard architecture for the task \citep{oh2015action, leibfried2016reward}. These models are then used to generate 100 samples of simulated experience for every interaction with the real environment. All three variations outperformed state-of-the-art Rainbow DQN in terms of data efficiency on 5 out of 6 games. Nevertheless, only the results of the online model are used for further discussion to ensure a fair comparison between the algorithms.

\citet{kaiser2019model} introduced Simulated Policy Learning (SimPLe). Similarly to the previous two architectures, it learns the model of the environment using a modified version of \citet{oh2015action}. It differs from previous approaches by employing PPO \citep{schulman2017ppo} as its RL agent and by using the learned model much more exhaustively. It uses the model similarly to \citet{holland2018dynadqn}, however it provides at least 800k samples of artificial data after every 6.4k interactions. The approach is then evaluated on a range of 26 different Atari games. It provides results that highly outperform both \citet{holland2018dynadqn} and \citet{azizzadenesheli2018gats} in terms of data efficiency achieving at least 2x improvement on over half of the games and more than 10x improvement on Freeway. To the best of our knowledge, SimPLe is the state of the art in terms of model-based data-efficient deep reinforcement learning; thus, it will be used as a primary model-based baseline throughout the rest of the paper.

However, model-based methods are not the only approaches for improving data efficiency of RL. Recently proposed Episodic Backward Update (EBU) \citep{lee2018ebu} showed that classical version of the DQN algorithm can be incrementally improved by using full episodes in the algorithm's replay mechanism and propagating rewards from end of the episode to its start, instead of sampling every step independently and uniformly at random.

\section{Data efficiency of standard approaches} \label{sect:eff}
We argue that classical DQN-like methods are not as data inefficient as they are often portrayed. They are simply used in a very inefficient way. Let us define ratio $\displaystyle r$ describing the number of training steps to the number of interactions with the environment. In the default setting $\tau_u = 4$. It means that the algorithm performs a single update of the network for every 4 interactions with the environment, i.e., $\displaystyle r = 1/4$. 

As explained in section 3, both the online-model-based algorithm from \citet{holland2018dynadqn} and SimPLe from \citet{kaiser2019model} first learn the approximated model of the environment. Then, this approximation is used to provide simulated samples of experience alongside the real data. Nevertheless, these samples, in the best case, can only provide as much real signal to the agent as was provided in the original data. However, as a byproduct of the agent's interactions with the learned model, the ratio $\displaystyle r$ significantly increases. \citet{holland2018dynadqn} performs 100 simulated steps for each real step causing $\displaystyle r = (1 + 100) * (1/4) = 25.25$. SimPLe executes 800k simulated steps after every 6.4k interactions with the real environment. Thus, if SimPLe was using DQN as its model-free component ratio $\displaystyle r$ would be even higher ($\displaystyle r = (800k + 6.4k) / 6.4k / 4 = 126 / 4 = 31.5$). 

The issue of inflated $\displaystyle r$ does not only affect model-based methods. Algorithm 2 from \citet{lee2018ebu} shows that model-free EBU algorithm performs it training step for every interaction with the environment. By itself it would increase defined above ratio to $\displaystyle r = 1$, however, it is not the end of the story. Single EBU update actually looks $\displaystyle T$ separate transitions where $\displaystyle T$ is the length of sampled episode so in the end we have $\displaystyle r = T >= 1$.

It seems unfair to allow novel methods to perform more training steps for each gathered data point without letting baselines to do the same. However, from the studies discussed above, only \citet{holland2018dynadqn} performed tests allowing DQN for extra updates\footnote{Their results showed that indeed model-based approach with the online model does not overperform model-free approach with extra updates. However, the study was mainly interested in thorough analysis, rather than improving the state of the art.}. GATS and EBU were compared solely to the standard version of DQN and SimPLe to the standard version of PPO algorithm together with the Rainbow DQN that, as stated in the paper, was hypertuned for sample efficiency (HRainbow). However, hyperparameters for HRainbow were not disclosed. We hypothesize, that the main reason behind improved data efficiency in the results is essentially increased $\displaystyle r$. 

\section{Experimental setup}

To test the above-mentioned hypothesis, we train both standard DQN as described in \citet{mnih2015dqn} and Rainbow DQN agent, as described in \citet{hessel2018rainbow}. Both with only a few small differences to increase ratio $\displaystyle r$. These slight modifications of the algorithms are referred to as Overtrained DQN (OTDQN) and Overtrained Rainbow (OTRainbow) throughout the rest of the paper. OTDQN is used for a fair evaluation of EBU, as EBU is just an incremental improvement over the pure DQN. Whereas we use OTRainbow for direct comparison with SimPLe because SimPLe claims state of the art results.

To create both OTDQN and OTRainbow we decrease period between updates as much as possible so $\displaystyle \tau_u = 1$ (thus $\displaystyle r = 1$). To ensure fainess with respect to EBU, we stick to this $\displaystyle r$ in case of OTDQN (EBU's $\displaystyle r >= 1$). However, for OTRainbow, we need to increase $\displaystyle r$ even further. Because it is impossible to do so using existing hyperparameters, we introduce a new parameter $k$ that specifies how many network updates should be performed every $\tau_u$ steps (similarly to DQN Extra Updates from \citet{holland2018dynadqn}). We find that $k = 8$ for Rainbow produces the best results (hence Rainbow's $\displaystyle r = 8$). We also decrease the epsilon decay period to 250K steps and 50K steps, target network update period 5000 and 500, and minimum replay history to 25K and 20K for OTDQN and OTRainbow respectively to make it compatible with low data settings.

Existing code from the Dopamine framework \citep{castro18dopamine} was modified, as explained above, to obtain overtrained algorithms. Dopamine was used for two reasons: (\textit{i}) it allows for quick and easy prototyping of new RL algorithms; (\textit{ii}) to ensure the same implementation for each version of the DQN (whether it is OTRainbow, HRainbow, or OTDQN). Both algorithms were then evaluated on different subsets of Atari games from the Atari Learning Environment. OTRainbow used the same 26 games as \citet{kaiser2019model}, whereas OTDQN used a random subset of 25 games from the whole pool of 49 games used by EBU.  We compare the outcomes to multiple different baselines: an agent that always chooses action uniformly at random (Random) and human score as reported by \citet{mnih2015dqn} (Human). Additionally, OTRainbow is compared directly with scores of SimPLe and HRainbow, as reported by \citet{kaiser2019model}. OTDQN, on the other hand, is analyzed against DQN with the hyperparameters from \citet{mnih2015dqn} (SDQN) and EBU scores reported in \citet{lee2018ebu}. 

To further ensure fairness of comparison, we choose to evaluate SimPLe and EBU in the same data settings they used in the original papers. The efficiency of OTRainbow and SimPLe is tested based on a mean score in the low data regime of 100k interactions with the real environment, while OTDQN and EBU use 2.5M interactions (10M frames). On top of that, we compare the overall performance of all models depending on the amount of available data using human normalized performance. I.e., we normalize agent scores on each game such that 0\% is the performance of the random agent, and 100\% corresponds to human score.

\section{Analysis}

\subsection{SimPLe}

\begin{figure}[h]
\begin{center}
\includegraphics[width=0.5\columnwidth]{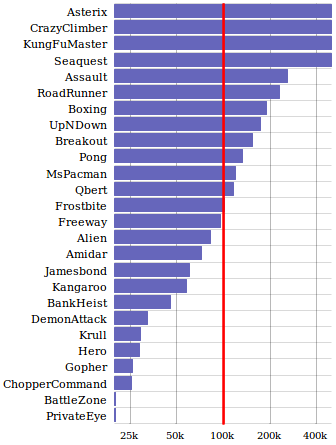}
\end{center}
\caption{Comparison of SimPLe with OTRainbow. Bars represent the number of interactions required by OTRainbow to reach the same score as SimPLe achieves using exactly 100k interactions. Notice logarithmic scale on X-axis.}
\label{fig:spvsotr}
\end{figure}

Overall, OTRainbow and SimPLe prove to be the best models evaluated in the 100k-interactions-only setting, without the clear winner between the two. Numerical results for this setting are shown in \tablename{ \ref{table:100k}}. Moreover \figurename{ \ref{fig:spvsotr}} compares OTRainbow and SimPLe directly, using graphical convention similar to \citet{kaiser2019model}. However, in this study, we use a logarithmic scale to denote the number of data samples needed to reach SimPLe's score. Doing so ensures that whether OTRainbow requires $\displaystyle n$ times more experience or $\displaystyle n$ times less, the absolute visual deviation from the SimPLe baseline is the same. Also, results are clipped to the absolute maximum deviation of 5x (i.e., 20k - 500k) as OTRainbow was evaluated on a maximum of 500k interactions due to computational constraints.

We can see that both OTRainbow and SimPLe outperform Random on all 26 games, surprisingly HRainbow did not manage to do the same. However, HRainbow falls behind Random only when playing Kangaroo. OTRainbow produces better scores than HRainbow on all games proving the weakness of HRainbow as a baseline. Interestingly, SimPLe's manages to beat HRainbow only on 20 out of 26 games. In terms of direct comparison between OTRainbow and SimPLe, they perform very evenly. OTRainbow outperforms SimPLe on exactly half of the games but is dominated by SimPLe on the remaining half. What is fascinating, however, is that the original paper behind SimPLe reported that efficiency on Freeway benefits most from the model-based approach, with SimPLe being 10x more efficient than HRainbow. This result is improved even further by OTRainbow as it managed to score over 8 points higher. It again shows that the improvement was rather an effect of an increased number of network updates than the model-based approach. We also calculate the human normalized performance for each algorithm. Full numerical results of these calculations can be seen in \tablename{ \ref{table:allnorm}} in Appendix \ref{app:results}. The mean human performance  However, the median human performance of OTRainbow beats SimPLe by over 10pp.  These results show that even the state-of-the-art model-based approach, highly tuned for achieving the best scores given a small number of interactions with the real environment, is not significantly more data-efficient than slightly modified existing techniques.

\begin{table}[h]
\caption{Mean scores produced by each approach in the low-data regime. Scores for OTRainbow, SimPLe, HRainbow, and Rainbow are obtained after 100k interactions with the real environment. Values in bold represent the top model for the game (ignores Human).}
\label{table:100k}
\begin{center}
\begin{tabular}{l|llllll}
\hline
Game           & OTRainbow & SimPLe & HRainbow & SRainbow & Human  & Random\\ \hline
Alien          & \textbf{824.7} & 616.9   & 290.6    & 318.7  & 6875 & 184.8 \\
Amidar         & \textbf{82.8}  & 74.3   & 20.8     & 32.5 & 1676 & 11.8  \\
Assault        & 351.9 & \textbf{527.2} & 285.7    & 231  & 1496 & 248.8 \\
Asterix        & 628.5  & \textbf{1128.3}  & 300.3 & 243.6  & 8503  & 233.7 \\
BankHeist      & \textbf{182.1}  & 34.2   & 34.5 & 15.55  & 734.4 & 15  \\
BattleZone     & \textbf{4060.6} & 4031.2 & 3363.5  & 3285.71 & 37800 & 2895\\
Boxing         & 2.5   & \textbf{7.8 }   & 0.9      & -24.8  & 4.3   & 0.3   \\
Breakout       & 9.84  & \textbf{16.4}   & 3.3      & 1.2  & 31.8  & 0.9   \\
ChopperCommand & \textbf{1033.33}  & 979.4  & 776.6  &  120  & 9882 & 671 \\
CrazyClimber   & 21327.8  & \textbf{62583.6} & 12558.3  &  2254.5 & 35411 & 7339\\
DemonAttack    & \textbf{711.8} & 208.1  & 431.6    & 163.6  & 3401 & 140 \\
Freeway        & \textbf{25}  & 16.7   & 0.1      & 0  & 29.6   & 0   \\
Frostbite      & 231.6 & \textbf{236.9}  & 140.1    & 60.2  & 4335  &  74 \\
Gopher         & \textbf{778}  & 596.8  & 748.3  &  431.2  & 2321 & 245.9\\
Hero           & \textbf{6458.8} & 2656.6 & 2676.3   & 487 & 25763& 224.6 \\
Jamesbond      & \textbf{112.3}  & 100.5  & 61.7     & 47.4  & 406.7  & 29.2  \\
Kangaroo       & \textbf{605.4} & 51.2   & 38.7     & 0  & 3035 & 42  \\
Krull          & \textbf{3277.9} & 2204.8 & 2978.8   & 1468 & 2395 & 1543.3\\
KungFuMaster   & 5722.2 & \textbf{14862.5} & 1019.4   & 0 & 22736& 616.5 \\
MsPacman       & 941.9 & \textbf{1480} & 364.3    & 67  & 15693 & 235.2 \\
Pong           & 1.3 & \textbf{12.8}   & -19.5  & -20.6 & 9.3   & -20.4 \\
PrivateEye     & \textbf{100}  & 35  & 42.1     &  0 & 69571 & 26.6  \\
Qbert          & 509.3  & \textbf{1288.8} & 235.6  &  123.46  & 13455 & 166.1 \\
RoadRunner     & 2696.7  & \textbf{5640.6} & 524.1    & 1588.46 & 7845 & 0   \\
Seaquest       & 286.92  & \textbf{683.3}  & 206.3    & 131.69  & 20182 & 61.1  \\
UpNDown        & 2847.6  & \textbf{3350.3} & 1346.3  & 504.6  & 9082 & 488.4
\end{tabular}
\end{center}
\end{table}

When comparing SimPLe to the variations of Rainbow DQN with respect to computational complexity, SimPLe is orders of magnitude more expensive. As shown in section \ref{sect:eff}, using SimPLe increases ratio $\displaystyle r$ 126 times, while the most computationally demanding variation of Rainbow - OTRainbow - increases $\displaystyle r$ 32 times. Thus, when taking into account only the reinforcement learning part, SimPLe already requires almost four times more network updates. On top of that, however, SimPLe has to perform expensive training of the world model. As reported by \citet{kaiser2019model}, a full version of SimPLe takes more than three weeks on 100k data points to complete the training. Using the same amount of data, OTRainbow can finish within the first 24 hours\footnote{When running on eight cores of Intel Haswell CPU.}.

\subsection{EBU}

\begin{figure}[h]
\begin{center}
\includegraphics[width=0.8\columnwidth]{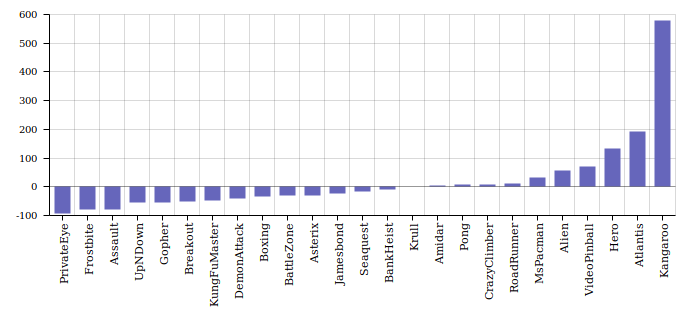}
\end{center}
\caption{Graph represent relative score of OTDQN compared to EBU in percents. Both are trained for 10M frames.}
\label{fig:ebu}
\end{figure}

Unlike SimPLe, EBU was evaluated in the data regime of 10M frames (2.5M steps). \tablename{ \ref{table:10m}} contains numerical results for all tested algorithms, namely EBU, OTDQN, and SDQN.  Because this setting allows for much more interactions than previously described low-data regime, none of the algorithms has a problem with surpassing the random agent. All of them have enough time to finish full exploration period and start converging to optimal policies. 

Similarly to SimPLe and OTRainbow, EBU and OTDQN compare evenly, with EBU having only a slight advantage. \citet{lee2018ebu} reported that EBU outperforms SDQN on 20 out of 25 games that are used in our experiments clearly being superior. However, when compared to OTDQN, it is better only on 14 games. What is more, it was stated in the original paper that although EBU does not surpass SDQN in all of the games, large improvements in Atlantis, Breakout, or VideoPinball offset these shortcomings. OTDQN highly outperforms EBU in both Atlantis and VideoPinball, showing that these scores were merely caused by an increased ratio $\displaystyle r$. It can be easily noticed in \figurename{ \ref{fig:ebu}} that visualizes relative performance of both algorithms. What is important to note, OTDQN still performs much less actual updates than EBU. Number of updates for both of the algorithms would be the same if and only if average episode length was $1$. In Atari setting, it's at least order of magnitude larger than that.

\begin{table}[h]
\caption{Mean scores produced by each approach. Scores for OTDQN, SDQN, and EBU, HRainbow, and Rainbow are obtained after 100k interactions with the real environment. Values in bold represent the top model for the game (ignores Human).}
\label{table:10m}
\begin{center}
\begin{tabular}{l|llllll}
\hline
Game           & OTQDN          & EBU            & SDQN     & Human  & Random\\ \hline
Alien          & \textbf{1288.4}& 894.2          & 690.3    & 6875   & 184.8 \\
Amidar         & \textbf{130.8} & 124.6          & 125.4    & 1676   & 11.8  \\
Assault        & 946.1          & \textbf{3677.0}& 2426.9   & 1496   & 248.8 \\
Asterix        & 1849.0        & 2533.2        &\textbf{2936.5}& 8503   & 233.7 \\
Atlantis       &\textbf{232010.0}& 87944.3       & 20666.8  & 29028  & 12850 \\
BankHeist      & 413.6          & \textbf{459.4} & 234.7    & 734.4  & 15  \\
BattleZone     & 17820.3        & \textbf{24748.5}& 22468.8 & 37800  & 2895\\
Boxing         & 48.8           &\textbf{72.7}   & 37.3     & 4.3    & 0.3   \\
Breakout       & 131.2          & \textbf{265.6} & 28.4     & 31.8   & 0.9   \\
CrazyClimber   &\textbf{101375.9}& 94135.0       & 74410.7  & 35411  & 7339\\
DemonAttack    & 4876.7         & \textbf{8368.2}& 7772.4   & 3401   & 140 \\
Frostbite      & 238.6          & \textbf{966.2} & 466.0    & 4335   & 74 \\
Gopher         & 1727.3         & \textbf{3634.7}& 1726.5   & 2321   & 245.9\\
Hero           & \textbf{7551.8}& 3398.6         & 2767.9   & 25763  & 224.6 \\
Jamesbond      & 408            & \textbf{519.5} & 183.4    & 406.7  & 29.2  \\
Kangaroo       & \textbf{4716.1}& 731.1          & 709.8    & 3035   & 42  \\
Krull          & 8774.7        & 8733.5        &\textbf{24109.1}& 2395   & 1543.3\\
KungFuMaster   & 13530.0        & \textbf{26069.7}& 21951.7 & 22736  & 616.5 \\
MsPacman       & \textbf{2080.5}& 1652.4         & 1861.8   & 15693  & 235.2 \\
Pong           & \textbf{19.0}  & 16.5           & -2.7     & 9.3    & -20.4 \\
PrivateEye     & 229.2          & \textbf{3610.0}& 1388.5   & 69571  & 26.6  \\
RoadRunner     & \textbf{17575.9}& 15681.5       & 8978.2   & 7845   & 0   \\
Seaquest       & 1583.6         & \textbf{1926.1}& 762.1    & 20182  & 61.1  \\
UpNDown        & 3195.6        & 6754.1        &\textbf{9468.0}& 9082   & 488.4 \\
VideoPinball   &\textbf{120703.0}& 78405.3       & 17803.7  & 17298  & 16257 

\end{tabular}
\end{center}
\end{table}

\section{Conclusion}
We presented an intuition why the previous research did not use fair baselines when comparing new advancements with currently existing methods. We suggested the way of using popular version of the DQN algorithm, namely OTDQN and OTRainbow, that leverage DQN's actual capabilities in terms of data efficiency. We experimentally proved that most of the recent advancement in model-free and model-based approaches show improved performance only due to an increase in ratio of the number of training updates to the number of interactions with the environment and not due to superiority of complicated and often extremely computationally expensive techniques that were proposed. In particular, it shows that the recent work in sample efficient deep reinforcement learning does not produce significant improvements over the existing methods upholding the position of model-free algorithms as the state of the art, both in terms of data efficiency and total performance, at least in visually-rich Atari domain. Through these results, we aim to underline the importance of using appropriate model-free baselines, such as OTRainbow, in the future research that tries to improve data efficiency of deep RL approaches. 

\bibliography{iclr2020_conference}
\bibliographystyle{iclr2020_conference}

\appendix
\section{Complete numerical results}
\label{app:results}

\begin{table}
\centering
\caption{Mean raw scores for each approach. Value in brackets after the name of the method indicates the number of training interactions performed before the evaluation.}
\label{table:all}
\resizebox{\columnwidth}{!}{%
\begin{tabular}{l|rrrrr} 
\hline
               & \multicolumn{1}{l}{OTRainbow (100k)} & \multicolumn{1}{l}{OTRainbow (500k)} & \multicolumn{1}{l}{SimPLe (100k)} & \multicolumn{1}{l}{HRainbow (100k)} & \multicolumn{1}{l}{SRainbow (100k)}  \\ 
\hline
Alien          & 824.7                                & 834.9                                & 616.9                             & 290.6                               & 318.7                                \\
Amidar         & 82.8                                 & 215.3                                & 74.3                              & 20.8                                & 32.5                                 \\
Assault        & 351.9                                & 549.3                                & 527.2                             & 285.7                               & 231                                  \\
Asterix        & 628.5                                & 930.9                                & 1128.3                            & 300.3                               & 243.6                                \\
BankHeist      & 182.1                                & 223.9                                & 34.2                              & 34.5                                & 15.5                                 \\
BattleZone     & 4060.6                               & 11093.8                              & 4031.2                            & 3363.5                              & 3285.7                               \\
Boxing         & 2.5                                  & 8.4                                  & 7.8                               & 0.9                                 & -24.8                                \\
Breakout       & 9.84                                 & 29.8                                 & 16.4                              & 3.3                                 & 1.2                                  \\
ChopperCommand & 1033.33                              & 1344                                 & 979.4                             & 776.6                               & 120                                  \\
CrazyClimber   & 21327.8                              & 28863.5                              & 62583.6                           & 12558.3                             & 2254.5                               \\
DemonAttack    & 711.8                                & 1303                                 & 208.1                             & 431.6                               & 163.6                                \\
Freeway        & 25                                   & 25.2                                 & 16.7                              & 0.1                                 & 0                                    \\
Frostbite      & 231.6                                & 255.6                                & 236.9                             & 140.1                               & 60.2                                 \\
Gopher         & 778                                  & 748.5                                & 596.8                             & 748.3                               & 431.2                                \\
Hero           & 6458.8                               & 12461.3                              & 2656.6                            & 2676.3                              & 487                                  \\
Jamesbond      & 112.3                                & 202.9                                & 100.5                             & 61.7                                & 47.4                                 \\
Kangaroo       & 605.4                                & 3398                                 & 51.2                              & 38.7                                & 0                                    \\
Krull          & 3277.9                               & 3718.1                               & 2204.8                            & 2978.8                              & 1468                                 \\
KungFuMaster   & 5722.2                               & 7261.7                               & 14862.5                           & 1019.4                              & 0                                    \\
MsPacman       & 941.9                                & 1803.1                               & 1480                              & 364.3                               & 67                                   \\
Pong           & 1.3                                  & 19.9                                 & 12.8                              & -19.5                               & -20.6                                \\
PrivateEye     & 100                                  & 100                                  & 35                                & 42.1                                & 0                                    \\
Qbert          & 509.3                                & 8346.2                               & 1288.8                            & 235.6                               & 123.4                                \\
RoadRunner     & 2696.7                               & 6887.5                               & 5640.6                            & 524.1                               & 1588.4                               \\
Seaquest       & 286.92                               & 323.9                                & 683.3                             & 206.3                               & 131.6                                \\
UpNDown        & 2847.6                               & 4067                                 & 3350.3                            & 1346.3                              & 504.6                                \\ 
\hline
               & \multicolumn{1}{l}{SRainbow (500k)}  & \multicolumn{1}{l}{SRainbow (1M)}    & \multicolumn{1}{l}{SRainbow (2M)} & \multicolumn{1}{l}{Human}           & \multicolumn{1}{l}{Random}           \\ 
\hline
Alien          & 481.5                                & 766.3                                & 1134.3                            & 6875                                & 184.8                                \\
Amidar         & 70.6                                 & 132.6                                & 249.2                             & 1676                                & 12                                   \\
Assault        & 468.6                                & 630.1                                & 1230.4                            & 1496                                & 249                                  \\
Asterix        & 352.6                                & 1038.7                               & 2320.1                            & 8503                                & 234                                  \\
BankHeist      & 17.5                                 & 304                                  & 872.1                             & 734.4                               & 15                                   \\
BattleZone     & 3346.3                               & 3453.7                               & 11894.8                           & 37800                               & 2895                                 \\
Boxing         & -29.5                                & 8.3                                  & 47.1                              & 4.3                                 & 0                                    \\
Breakout       & 4.5                                  & 15.6                                 & 32.4                              & 31.8                                & 1                                    \\
ChopperCommand & 433.5                                & 915.6                                & 1810.1                            & 9882                                & 671                                  \\
CrazyClimber   & 26090.9                              & 66577.2                              & 98461.7                           & 35411                               & 7339                                 \\
DemonAttack    & 213.6                                & 487.8                                & 1748                              & 3401                                & 140                                  \\
Freeway        & 8.2                                  & 27.45                                & 31.9                              & 29.6                                & 0                                    \\
Frostbite      & 275.2                                & 512.3                                & 2408.9                            & 4335                                & 74                                   \\
Gopher         & 426.6                                & 2119.2                               & 3649.9                            & 2321                                & 246                                  \\
Hero           & 326.6                                & 3216                                 & 7875                              & 25763                               & 225                                  \\
Jamesbond      & 50.2                                 & 236.1                                & 472.2                             & 406.7                               & 29                                   \\
Kangaroo       & 153.7                                & 567.4                                & 4252.9                            & 3035                                & 42                                   \\
Krull          & 4714.2                               & 6187.9                               & 6136                              & 2395                                & 1543                                 \\
KungFuMaster   & 596.7                                & 10544.3                              & 16284.5                           & 22736                               & 617                                  \\
MsPacman       & 1244.2                               & 1918.6                               & 2301.5                            & 15693                               & 235                                  \\
Pong           & -20.6                                & -16.5                                & 10.6                              & 9.3                                 & -20                                  \\
PrivateEye     & 692.8                                & 169.1                                & 92.5                              & 69571                               & 27                                   \\
Qbert          & 450.6                                & 1189                                 & 4046.9                            & 13455                               & 166                                  \\
RoadRunner     & 1261.9                               & 13793.9                              & 31159                             & 7845                                & 0                                    \\
Seaquest       & 181.2                                & 378.4                                & 1496.5                            & 20182                               & 61                                   \\
UpNDown        & 1284.6                               & 5566.3                               & 10298.7                           & 9082                                & 488.4                               
\end{tabular}
}
\end{table}   

\begin{table}
\centering
\caption{Mean human normalized score for each approach. Value in brackets after the name of the method indicates the number of training interactions performed before the evaluation.}
\label{table:allnorm}
\resizebox{\columnwidth}{!}{%
\begin{tabular}{l|rrrr} 
\hline
               & \multicolumn{1}{l}{OTRainbow (100k)} & \multicolumn{1}{l}{OTRainbow (500k)} & \multicolumn{1}{l}{SimPLe (100k)} & \multicolumn{1}{l}{HRainbow (100k)}  \\ 
\hline
Alien          & 9.56\%                               & 9.72\%                               & 6.46\%                            & 1.58\%                               \\
Amidar         & 4.27\%                               & 12.23\%                              & 3.76\%                            & 0.54\%                               \\
Assault        & 8.27\%                               & 24.09\%                              & 22.32\%                           & 2.96\%                               \\
Asterix        & 4.77\%                               & 8.43\%                               & 10.82\%                           & 0.81\%                               \\
BankHeist      & 23.23\%                              & 29.04\%                              & 2.67\%                            & 2.71\%                               \\
BattleZone     & 3.34\%                               & 23.49\%                              & 3.26\%                            & 1.34\%                               \\
Boxing         & 55.00\%                              & 202.50\%                             & 187.50\%                          & 15.00\%                              \\
Breakout       & 28.93\%                              & 93.53\%                              & 50.16\%                           & 7.77\%                               \\
ChopperCommand & 3.93\%                               & 7.31\%                               & 3.35\%                            & 1.15\%                               \\
CrazyClimber   & 49.83\%                              & 76.68\%                              & 196.80\%                          & 18.59\%                              \\
DemonAttack    & 17.53\%                              & 35.66\%                              & 2.09\%                            & 8.94\%                               \\
Freeway        & 84.46\%                              & 85.14\%                              & 56.42\%                           & 0.34\%                               \\
Frostbite      & 3.70\%                               & 4.26\%                               & 3.82\%                            & 1.55\%                               \\
Gopher         & 25.64\%                              & 24.22\%                              & 16.91\%                           & 24.21\%                              \\
Hero           & 24.41\%                              & 47.91\%                              & 9.52\%                            & 9.60\%                               \\
Jamesbond      & 22.01\%                              & 46.01\%                              & 18.89\%                           & 8.61\%                               \\
Kangaroo       & 18.82\%                              & 112.13\%                             & 0.31\%                            & -0.11\%                              \\
Krull          & 203.66\%                             & 255.35\%                             & 77.67\%                           & 168.55\%                             \\
KungFuMaster   & 23.08\%                              & 30.04\%                              & 64.40\%                           & 1.82\%                               \\
MsPacman       & 4.57\%                               & 10.14\%                              & 8.05\%                            & 0.84\%                               \\
Pong           & 73.06\%                              & 135.69\%                             & 111.78\%                          & 3.03\%                               \\
PrivateEye     & 0.11\%                               & 0.11\%                               & 0.01\%                            & 0.02\%                               \\
Qbert          & 2.58\%                               & 61.56\%                              & 8.45\%                            & 0.52\%                               \\
RoadRunner     & 34.37\%                              & 87.79\%                              & 71.90\%                           & 6.68\%                               \\
Seaquest       & 1.12\%                               & 1.31\%                               & 3.09\%                            & 0.72\%                               \\
UpNDown        & 27.45\%                              & 41.64\%                              & 33.30\%                           & 9.98\%                               \\
\textbf{Median}         & \textbf{20.42\%}                              & \textbf{32.85\%  }                            & \textbf{10.17\%}                           & \textbf{2.27\%   }                            \\ 
\hline
               & \multicolumn{1}{l}{SRainbow (100k)}  & \multicolumn{1}{l}{SRainbow (500k)}  & \multicolumn{1}{l}{SRainbow (1M)} & \multicolumn{1}{l}{SRainbow (2M)}    \\ 
\hline
Alien          & 2.00\%                               & 4.43\%                               & 8.69\%                            & 14.19\%                              \\
Amidar         & 1.24\%                               & 3.53\%                               & 7.26\%                            & 14.27\%                              \\
Assault        & -1.43\%                              & 17.62\%                              & 30.57\%                           & 78.70\%                              \\
Asterix        & 0.12\%                               & 1.44\%                               & 9.73\%                            & 25.23\%                              \\
BankHeist      & 0.07\%                               & 0.35\%                               & 40.17\%                           & 119.14\%                             \\
BattleZone     & 1.12\%                               & 1.29\%                               & 1.60\%                            & 25.78\%                              \\
Boxing         & -627.50\%                            & -745.00\%                            & 200.00\%                          & 1170.00\%                            \\
Breakout       & 0.97\%                               & 11.65\%                              & 47.57\%                           & 101.94\%                             \\
ChopperCommand & -5.98\%                              & -2.58\%                              & 2.66\%                            & 12.37\%                              \\
CrazyClimber   & -18.11\%                             & 66.80\%                              & 211.02\%                          & 324.60\%                             \\
DemonAttack    & 0.72\%                               & 2.26\%                               & 10.67\%                           & 49.31\%                              \\
Freeway        & 0.00\%                               & 27.70\%                              & 92.74\%                           & 107.77\%                             \\
Frostbite      & -0.32\%                              & 4.72\%                               & 10.29\%                           & 54.80\%                              \\
Gopher         & 8.93\%                               & 8.71\%                               & 90.28\%                           & 164.04\%                             \\
Hero           & 1.03\%                               & 0.40\%                               & 11.71\%                           & 29.96\%                              \\
Jamesbond      & 4.82\%                               & 5.56\%                               & 54.81\%                           & 117.35\%                             \\
Kangaroo       & -1.40\%                              & 3.73\%                               & 17.55\%                           & 140.69\%                             \\
Krull          & -8.84\%                              & 372.30\%                             & 545.33\%                          & 539.24\%                             \\
KungFuMaster   & -2.79\%                              & -0.09\%                              & 44.88\%                           & 70.83\%                              \\
MsPacman       & -1.09\%                              & 6.53\%                               & 10.89\%                           & 13.37\%                              \\
Pong           & -0.67\%                              & -0.67\%                              & 13.13\%                           & 104.38\%                             \\
PrivateEye     & -0.04\%                              & 0.96\%                               & 0.20\%                            & 0.09\%                               \\
Qbert          & -0.32\%                              & 2.14\%                               & 7.70\%                            & 29.20\%                              \\
RoadRunner     & 20.25\%                              & 16.09\%                              & 175.83\%                          & 397.18\%                             \\
Seaquest       & 0.35\%                               & 0.60\%                               & 1.58\%                            & 7.13\%                               \\
UpNDown        & 0.19\%                               & 9.27\%                               & 59.09\%                           & 114.16\%                             \\
\textbf{Median}         &\textbf{ 0.03\%}                               & \textbf{3.63\% }                              & \textbf{15.34\%}                           & \textbf{74.77\% }   \\

\end{tabular}
}
\end{table}   

\end{document}

%% file: math_commands.tex
\usepackage{amsmath,amsfonts,bm}

\def\eqref#1{equation~\ref{#1}}

\def\1{\bm{1}}

\DeclareMathAlphabet{\mathsfit}{\encodingdefault}{\sfdefault}{m}{sl}
\SetMathAlphabet{\mathsfit}{bold}{\encodingdefault}{\sfdefault}{bx}{n}

\def\sA{{\mathbb{A}}}

\def\sR{{\mathbb{R}}}
\def\sS{{\mathbb{S}}}

